\documentclass[11pt,a4paper]{article}
\usepackage[hyperref]{acl2020}
\usepackage{times}
\usepackage{latexsym}
\usepackage{graphicx}
\usepackage{booktabs}
\usepackage[disable]{todonotes}

\usepackage{amsmath,amssymb}
\usepackage{dirtytalk}
\usepackage{microtype}

\aclfinalcopy 

\newcommand{\note}[4][]{\todo[author=#2,color=#3,size=\scriptsize,fancyline,caption={},#1]{#4}} 
\newcommand{\mans}[2][]{\note[#1]{mans}{orange!40}{#2}}

\title{The SIGMORPHON 2020 Shared Task on\\Unsupervised Morphological Paradigm Completion}

\author{
Katharina Kann\thanks{~~Equal contribution.} \\
University of Colorado Boulder \\
\texttt{katharina.kann@colorado.edu}
\And 
Arya D. McCarthy\textsuperscript{*}  \\
Johns Hopkins University \\
\texttt{arya@jhu.edu} \\\AND Garrett Nicolai \\
University of British Columbia \\
\texttt{garrett.nicolai@ubc.ca}\And Mans Hulden \\
University of Colorado Boulder \\
\texttt{mans.hulden@colorado.edu}
}

\date{}

\begin{document}
\maketitle
\begin{abstract}
In this paper, we describe the findings of the SIGMORPHON 2020 shared task on unsupervised morphological paradigm completion (SIGMORPHON 2020 Task 2), a novel task in the field of inflectional morphology. Participants were asked to submit systems which take raw text and a list of lemmas as input, and output all inflected forms, i.e., the entire morphological paradigm, of each lemma.
In order to simulate a realistic use case, we first released data for 5 development languages. However, systems were officially evaluated on 9 surprise languages, which were only revealed a few days before the  submission deadline.
We provided a modular baseline system, which is a pipeline of 4 components.
3 teams submitted a total of 7 systems, but, surprisingly, none of the submitted systems was able to improve over the baseline on average over all 9 test languages. Only on 3 languages did a submitted system obtain the best results.
This shows that unsupervised morphological paradigm completion is still largely unsolved. We present an analysis here, so that this shared task will ground further research on the topic.
\end{abstract}

\section{Introduction}
In morphologically rich languages, words \textit{inflect}: grammatical information like person, number, tense, and case are incorporated into the word itself, rather than expressed via function words. Not all languages mark the same properties: German nouns, for instance, have more inflected forms than their English counterparts.

When acquiring a language, humans usually learn to inflect words without explicit instruction. Thus, most native speakers are capable of generating inflected forms even of artificial lemmas \cite{berko1958child}.
However, 
models that can generate paradigms without explicit morphological training have not yet been developed.
We anticipate that such systems will be extremely useful, as they will open the possibility of rapid development of first-pass inflectional paradigms in a large set of languages. These can be utilized both \emph{in se} for generation and as a starting point for elicitation \citep{sylak-glassman-etal-2016-remote}, thus aiding
the development of low-resource human language technologies \citep{lorelei}.

\begin{figure}[t]
    \centering
    \includegraphics[width=.93\columnwidth]{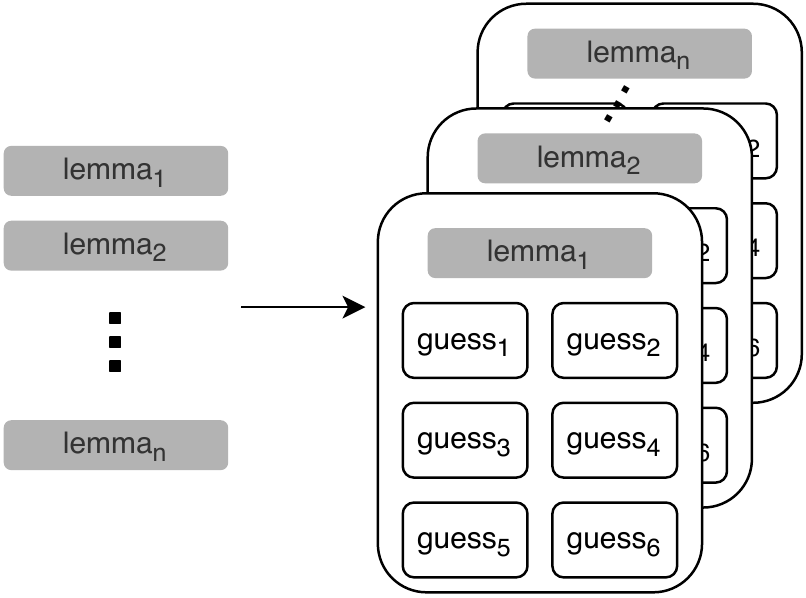}
    \caption{The task of \emph{unsupervised morphological paradigm completion} \citep{jin2020unsupervised} consists of generating complete inflectional paradigms for given lemmas, with the only additional available information being a corpus without annotations. 
    }
    \label{fig:task}
\end{figure}
In this paper, we present the SIGMORPHON 2020 shared task on unsupervised morphological paradigm completion (SIGMORPHON 2020 Task 2). 
We asked participants to produce systems that can learn to inflect \textit{in an unsupervised fashion}: 
given a small corpus (the Bible) together with a list of lemmas for each language, systems for the shared task should output all corresponding inflected forms. 
In their output, systems had to mark which forms expressed the same morphosyntactic features, e.g., demonstrate knowledge of the fact that \textit{walks} is to \textit{walk} as \textit{listens} is to \textit{listen}, despite not recognizing the morphological features explicitly.
We show a visualization of our shared task setup in \autoref{fig:task}.

Unsupervised morphological paradigm completion requires solving multiple subproblems either explicitly or implicitly. First, a system needs to figure out which words in the corpus belong to the same paradigm. This can, for instance, be done via string similarity: \textit{walks} is similar to \textit{walk}, but less so to \textit{listen}. Second, it needs to figure out the shape of the paradigm.
This requires detecting which forms of different lemmas express the same morphosyntactic features, even if they are not constructed from their respective lemmas in the exact same way. Third, a system needs to generate all forms not attested in the provided corpus. Using the collected inflected forms as training data, this can be reduced to the supervised morphological inflection task \cite{cotterell-etal-2016-sigmorphon}.

This year's submitted systems can be split into two categories: those that built on the baseline (\textbf{Retrieval+X}) and those that did not (\textbf{Segment+Conquer}). The baseline system is set up as a pipeline which performs the following steps: edit tree retrieval, additional lemma retrieval, paradigm size discovery, and inflection generation \cite{jin2020unsupervised}. As it is highly modular, we provided two versions that employ different inflection models.\footnote{In this report, we use the words \textit{baseline} and \textit{baselines} interchangeably.}
All systems built on the baseline substituted the morphological inflection component.

No system outperformed the baseline overall. However, two \textbf{Retrieval+X} models  slightly improved over the baseline on three individual languages. We conclude that the task of unsupervised morphological paradigm completion is still an open challenge,
and we hope that this shared task will inspire future research in this area.

\section{Task and Evaluation}
\subsection{Unsupervised Morphological Paradigm Completion}
\paragraph{Informal description.}
The task of unsupervised morphological paradigm completion mimics a setting where the only resources available in a language are a corpus and a short list of dictionary forms, i.e., lemmas. The latter could, for instance, be obtained via basic word-to-word translation. The goal is to generate all inflected forms of the given lemmas.

For an English example, assume the following lemma list to be given:
\begin{align*}
    &walk \\\nonumber
    &listen \nonumber
\end{align*}
\normalsize{
With the help of raw text, systems should then produce an output like this:}
\begin{align}\nonumber
    &\mathit{walk} ~ \mathit{walk} ~ 1\\\nonumber
    &\mathit{walk} ~ \mathit{walks} ~ 2 \\\nonumber
    &\mathit{walk} ~ \mathit{walked} ~ 3 \\\nonumber
    &\mathit{walk} ~ \mathit{walking} ~ 4 \\\label{al:1}
    &\mathit{walk} ~ \mathit{walked} ~ 5 \\\nonumber
    &\mathit{listen} ~ \mathit{listens} ~ 2 \\\nonumber
    &\mathit{listen} ~ \mathit{listened} ~ 5 \\\nonumber
    &\mathit{listen} ~ \mathit{listened} ~ 3 \\\nonumber
    &\mathit{listen} ~ \mathit{listening} ~ 4 \\\nonumber
    &\mathit{listen} ~ \mathit{listen} ~ 1 \nonumber
\end{align}
\normalsize 
The numbers serve as unique identifiers for paradigm slots: in above example, "4" corresponds to the \textit{present participle}. The inflections \textit{walking} and \textit{talking} therefore belong to the same paradigm slot. For the task, participants are not provided any knowledge of the grammatical content of the slots.\mans[disable]{This could be clarified: do we mean participants don't ever have any knowledge of the grammatical content of the slots, or something like ``For the purposes of the task, the actual linguistic definition of the slots is irrelevant.''}

\paragraph{Formal definition.} We denote the paradigm $\pi(\ell)$ of a lemma $\ell$ as
\begin{equation}
    \pi(\ell) = \left\langle f(\ell, \vec{t}_\gamma)\right\rangle_{\gamma \in \Gamma(\ell)}\text{,}
\end{equation}
with $f : \Sigma^* \times \mathcal{T} \to \Sigma^*$ being a function that maps a lemma and a vector of morphological features~$\vec{t}_\gamma \in \mathcal{T}$ expressed by paradigm slot $\gamma$ to the corresponding inflected form. 
$\Gamma(\ell)$ is the set of slots in lemma $\ell$'s paradigm.

We then formally describe the task of unsupervised morphological paradigm completion as follows.
Given a corpus $\mathcal{D}=w_1,\dots,w_{|\mathcal{D}|}$ 
together with a list $\mathcal{L} = \{\ell_j\}$ 
of $|\mathcal{L}|$~lemmas belonging to the same part of speech,\footnote{This edition of the shared task was only concerned with verbs, though we are considering extending the task to other parts of speech in the future.} 
unsupervised morphological paradigm completion consists of
generating the paradigms~$\{\pi(\ell)\}$ 
of all lemmas $\ell \in \mathcal{L}$.

\paragraph{Remarks. } It is impossible for unsupervised systems to predict the names of the features expressed by paradigm slots, an arbitrary decision made by human annotators. This is why, for the shared task, we asked systems to mark which forms belong to the same slot by numbering them, e.g., to predict that \textit{walked} is the form for slot 3, while \textit{listens} corresponds to slot 2.

\subsection{Macro-averaged Best-Match Accuracy}
The official evaluation metric was macro-averaged best-match accuracy \citep[BMAcc;][]{jin2020unsupervised}.

In contrast to supervised morphological inflection \cite{cotterell-etal-2016-sigmorphon}, our task cannot be evaluated with word-level accuracy. For the former, one can compare the prediction for each lemma and morphological feature vector to the ground truth. However, for unsupervised  paradigm completion, this requires a mapping from predicted slots to the gold standard's paradigm slots. 

BMAcc, thus, first computes the word-level accuracy each predicted slot would obtain against each true slot. It then constructs a complete bipartite graph, with those accuracies as edge weights. This enables computing of the maximum-weight full matching with the algorithm of \citet{karp1980algorithm}. BMAcc then corresponds to the sum of all accuracies for the best matching, divided by the maximum of the number of gold and predicted slots.

BMAcc penalizes systems for predicting a wrong number of paradigm slots. However, detecting the correct number of \textit{identical} slots -- something we encounter in some languages due to syncretism -- is extremely challenging. Thus, we merge slots with identical forms for all lemmas in both the predictions and the ground truth before evaluating.

\paragraph{Example.} Assume our gold standard is (\ref{al:1}) (the complete, 5-slot English paradigms for the verbs \emph{walk} and \emph{listen}) and a system outputs the following, including an error in the fourth row:
\begin{align*}
    &\mathit{walk} ~ \mathit{walks} ~ 1 \\\nonumber
    &\mathit{walk} ~ \mathit{walking} ~ 2 \\\nonumber
    &\mathit{listen} ~ \mathit{listens} ~ 1 \\\nonumber
    &\mathit{listen} ~ \mathit{listenen} ~ 2 \nonumber
\end{align*}
First, we merge slots 3 and 5 in the gold standard, since they are identical for both lemmas. Ignoring slot 5, we then compute the BMAcc as follows.
Slot 1 yields an accuracy of 100\% as compared to gold slot 2, and 0\% otherwise. Similarly, slot 2 reaches an accuracy of 50\% for gold slot 4, and 0\% otherwise.\mans{Like this example, but could (1) be reproduced here side-by-side so the reader doesn't have to look back to remember what slot 4 was?} Additionally, given the best mapping of those two slots, we obtain 0\% accuracy for gold slots 1 and 3. Thus, the BMAcc is 
\begin{equation}
    \textrm{BMAcc} = \frac{1 + 0.5 + 0 + 0}{4} = 0.375
\end{equation}

\section{Shared Task Data}
\subsection{Provided Resources}
We provided data for 5 development and 9 test languages. The development languages were available for system development and hyperparameter tuning, while the test languages were released shortly before the shared task deadline.
For the test languages, no ground truth data was available before system submission. This setup emulated a real-world scenario with the goal to create a system for languages about which we have no information. 

For the raw text corpora, we leveraged the JHU Bible Corpus \cite{mccarthy-etal-2020-johns}. This resource covers 1600 languages, which will enable future work to quickly produce systems for a large set of languages. Additionally, using the Bible allowed for a fair comparison of models across languages without potential confounds such as domain mismatch. 7 of the languages have only the New Testament available (approximately 8k sentences), and 7 have both the New and Old Testaments (approximately 31k sentences). 

\begin{table*}
    \centering
    \small
    \setlength{\tabcolsep}{5.5pt}
    \begin{tabular}{r | l|rrrrr}
    \toprule
   && MLT & FAS & POR & RUS & SWE \\ 
   \midrule
1 & \# Tokens in corpus  & 193257 & 227584 & 828861 & 727630 & 871707 \\
2 & \# Types in corpus  & 16017 & 11877 & 31446 & 46202 & 25913 \\
3 & \# Lemmas  & 20 & 100 & 100 & 100 & 100 \\
4 & \# Lemmas in corpus  & 10 & 22 & 50 & 50 & 50 \\
5 & \# Inflections  & 640 & 13600 & 7600 & 1600 & 1100 \\
6 & \# Inflections in corpus  & 252 & 545 & 1037 & 306 & 276 \\
7 & Paradigm size  & 16 & 136 & 76 & 16 & 11 \\
8 & Paradigm size (merged)  & 15 & 132 & 59 & 16 & 11 \\
    \bottomrule
    \end{tabular}
    \caption{Dataset statistics: \textbf{development} languages. \# Inflections=number of inflected forms in the gold file, token-based; \# Inflections in corpus=number of inflections from the gold file which can be found in the corpus, token-based; Paradigm size=number of different morphological feature vectors in the dataset for the language; Paradigm size (merged)=paradigm size, but counting slots with all forms being identical only once.}
    \label{tab:dev_data}
\end{table*}
\begin{table*}
    \centering
    \small
    \setlength{\tabcolsep}{5.5pt}
    \begin{tabular}{r | l|rrrrrrrrr}
    \toprule
   && EUS & BUL & ENG & FIN & DEU & KAN & NAV & SPA & TUR \\ 
   \midrule
1 & \# Tokens in corpus  & 195459 & 801657 & 236465 & 685699 & 826119 & 193213 & 104631 & 251581 & 616418 \\
2 & \# Types in corpus  & 18367 & 37048 & 7144 & 54635 & 22584 & 28561 & 18799 & 9755 & 59458 \\
3 & \# Lemmas  & 20 & 100 & 100 & 100 & 100 & 20 & 100 & 100 & 100 \\
4 & \# Lemmas in corpus  & 4 & 50 & 50 & 50 & 50 & 10 & 9 & 50 & 50 \\
5 & \# Inflections  & 10446 & 5600 & 500 & 14100 & 2900 & 2612 & 3000 & 7000 & 12000 \\
6 & \# Inflections in corpus  & 97 & 915 & 127 & 497 & 631 & 1040 & 54 & 630 & 986 \\
7 & Paradigm size  & 1659 & 56 & 5 & 141 & 29 & 85 & 30 & 70 & 120 \\
8 & Paradigm size (merged)  & 1658 & 54 & 5 & 141 & 20 & 59 & 30 & 70 & 120 \\
    \bottomrule
    \end{tabular}
    \caption{Dataset statistics: \textbf{test} languages. \# Inflections=number of inflected forms in the gold file, token-based; \# Inflections in corpus=number of inflections from the gold file which can be found in the corpus, token-based; Paradigm size=number of different morphological feature vectors in the dataset for the language; Paradigm size (merged)=paradigm size, but counting slots with all forms being identical only once.}
    \label{tab:test_data}
\end{table*}

All morphological information 
was taken from UniMorph \citep{sylak-glassman-etal-2015-language,kirov-etal-2018-unimorph}, a resource which contains paradigms for more than 100 languages. However, this information was only accessible to the participants for the development languages. UniMorph paradigms were further used internally for evaluation on the test languages---this data was then released after the conclusion of the shared task.

\subsection{Languages}
During the development phase of the shared task, we released 5 languages to allow participants
to investigate various design decisions: Maltese ({MLT}), Persian ({FAS}), 
Portuguese ({POR}), Russian ({RUS}), and Swedish ({SWE}).
These languages are typologically and genetically varied, 
representing a number of verbal inflectional phenomena.
Swedish and Portuguese are typical of Western European languages, and mostly
exhibit fusional, suffixing verbal inflection.  Russian, as an exemplar of 
Slavic languages, is still mostly suffixing, but does observe regular ablaut,
and has considerable phonologically-conditioned allomorphy.
Maltese is a Semitic language with a heavy Romance influence, and verbs
combine templatic and suffixing inflection.  Persian is mostly suffixing,
but does allow for verbal inflectional prefixation, such as negation and marking subjunctive mood.  Since the development languages were used for system tuning, their scores did not count towards
the final ranking.

After a suitable period for system development and tuning, we released nine 
test languages:
Basque ({EUS}), Bulgarian ({BUL}), English ({ENG}), Finnish ({FIN}), German 
({DEU}), Kannada ({KAN}), Navajo ({NAV}), Spanish ({SPA}), and Turkish ({TUR}).
Although these languages observe many features common to the development 
languages, such as fusional inflection, suffixation, and ablaut, they also cover
inflectional categories absent in the development languages.
Navajo, unlike any of the development languages, is strongly prefixing.  Basque, 
Finnish, and Turkish are largely agglutinative, with long, complex affix chains
that are difficult to identify through longest suffix matching. Furthermore,
Finnish and Turkish feature vowel harmony and consonant gradation, which both require a method to identify allomorphs correctly to be able to merge different variants of the same paradigm slot.

\begin{table*}
    \centering
    \small
    \setlength{\tabcolsep}{8.pt}
    \begin{tabular}{l l l l}
    \toprule
    Institution & Systems & Rank & Description Paper \\ \midrule
KU-CST & KU-CST-1 & 7 & \citet{Agirrezabal2020} \\
KU-CST & KU-CST-2 & 6 & \citet{Agirrezabal2020} \\ \midrule
IMS-CUBoulder & IMS-CUBoulder-1 & 5 & \citet{mager-kann-2020} \\
IMS-CUBoulder & IMS-CUBoulder-2 & 1 & \citet{mager-kann-2020} \\ \midrule
NYU-CUBoulder & NYU-CUBoulder-1 & 4 & \citet{singer-kann-2020} \\
NYU-CUBoulder & NYU-CUBoulder-2 & 2 & \citet{singer-kann-2020} \\
NYU-CUBoulder & NYU-CUBoulder-3 & 3 & \citet{singer-kann-2020} \\ \bottomrule
    \end{tabular}
    \caption{All submitted systems by institution, together with a reference to their description paper. The rank is relative to all other submitted systems and does not take the baselines into account.}
    \label{tab:submissions}
\end{table*}

\begin{figure*}[h]
    \centering
    \includegraphics[width=.88\linewidth]{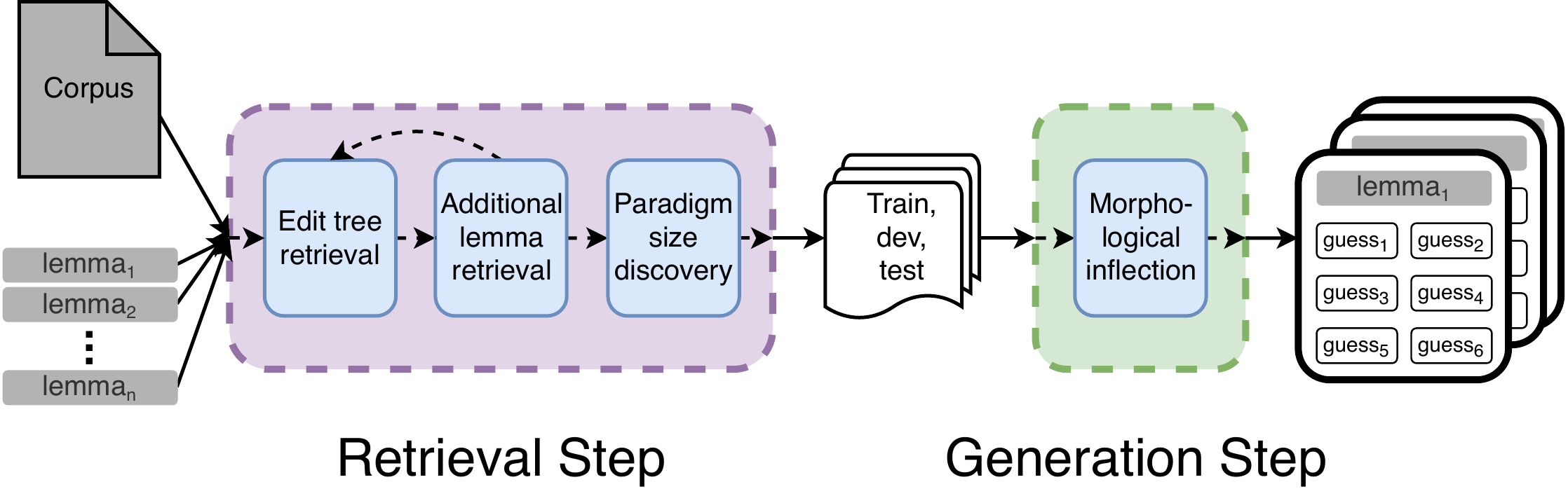}
    \caption{Our baseline system: the retrieval component bootstraps lemma--form--slot triplets, which are then used by the generation component to generate unobserved inflections in the paradigm of each input lemma.}
    \label{fig:my_label}
\end{figure*}

\subsection{Statistics} 
Statistics of the resources provided for all languages are shown in \autoref{tab:dev_data} for the development languages and in \autoref{tab:test_data} for the test languages.

The token count (line 1) and, thus, the size of the provided Bible corpora, differs between 104,631 (Kannada) and 871,707 (Swedish). This number depends both on the typology of a language and on the completeness of the provided Bible translation. 
The number of types (line 2) is between 7,144 (English) and 59,458 (Turkish). It is strongly influenced by how morphologically rich a language is, i.e., how large the paradigms are, which is often approximated with the \emph{type--token ratio}. The verbal paradigm size is listed in line 7: English has  with a size of 5 the smallest paradigms, and, correspondingly, the lowest type count. Turkish, which has the highest number of types, in contrast, has large paradigms (120). The last line serves as an indicator of syncretism: subtracting line 8 from line 7 results in the number of paradigm slots that have been merged as a language evolved to use identical forms for different inflectional categories.

Lines 3 and 4 show the number of lemmas in the lemma lists for all languages, as well as the number of lemmas which can be found in the corpus. For the majority of languages, 100 lemmas are provided, out of which 50 appear in the Bible. Exceptions are Maltese (20, 10), Persian (100, 22), Basque (20, 4), Kannada (20, 10), and Navajo (100, 9). These are due to limited UniMorph coverage.

In line 5, we list the number of total inflections, counting each one in the case of identical forms, i.e., this corresponds to the number of lines in our gold inflection file. English, due to its small verbal paradigm size, has only 500 inflections in our data. Conversely, Finnish has with 14,100 the largest number of inflections. Line 6 describes how many of the forms from line 5 appear in the corpus. As before, all forms are counted, even if they are identical. For all languages, a large majority of forms cannot be found in the corpus. This makes the task of unsupervised morphological paradigm completion with our provided data a challenging one.

\section{Systems}
In this section, we first review the baseline before describing the submitted systems. An additional overview of the submissions is shown in \autoref{tab:submissions}.

\subsection{Baseline}
We compared all submissions to the baseline system of \citet{jin2020unsupervised}, graphically summarized in \autoref{fig:my_label}. 
It is a pipeline system, which consists of 4 separate modules, which, in turn, can be grouped into two major components: \textit{retrieval} and \textit{generation}. The \textit{retrieval component} discovers and returns inflected forms -- and, less importantly, additional lemmas -- from the provided Bible corpus. The \textit{generation component}  produces new inflected forms which cannot be found in the raw text.

The \textbf{retrieval component} performs three steps: First, it extracts the most common edit trees \cite{chrupala2008towards}, i.e., it detects regularities with regards to word formation, based on the lemma list. If, for instance,  both \textit{walk} and \textit{listen} are the lemmas provided and both \textit{walked} and \textit{listened} are encountered in the corpus, the system notes that appending \textit{-ed} is a common transformation, which might correspond to an inflectional strategy. 

Second, it retrieves new lemmas, with the goal to gather additional evidence for our collected edit trees. If, for instance, it has already identified the suffix \textit{-ed} as an inflectional marker, finding both \textit{pray} and \textit{prayed} in the Bible is an indication that \textit{pray} might be a lemma. New lemmas can then, in turn, be used to detect new regularities, e.g., in the case that \textit{listen} and \textit{listens} as well as \textit{pray} and \textit{prays} are attested in the corpus, but \textit{walks} is not. 
Due to their complementary nature, components one and two can, as a unit, be applied iteratively to bootstrap a larger list of lemmas and transformations. For the baseline, we apply each of them only once.

Finally, the baseline's retrieval component predicts the paradigm size by analyzing which edit trees might be representing the same inflection. For instance, the suffixes \textit{-d} and \textit{-ed} both represent the past tense in English.
The output of the retrieval component is a list of inflected forms with their lemmas, annotated with a paradigm slot number.

The \textbf{generation component} receives this output and prepares the data to train an inflectional generator. First, identified inflections are divided into a training and development split, and missing paradigm slots are identified. The generator is trained on the discovered inflections, and new forms are predicted for each missing slot.

We used two morphological inflection systems for the two variants of our baseline: the non-neural baseline from \citet{cotterell-etal-2017-conll} and the model proposed by \citet{makarov-clematide-2018-imitation}. Both are highly suitable for the low-resource setting.

\subsection{Submitted Systems: Retrieval+X}
We now describe the first category of shared task submissions: Retrieval+X. Systems in this category leverage the retrieval component of the baseline, while substituting the morphological inflection component with a custom inflection system. 

The \textbf{IMS--CUBoulder team} relied on LSTM \cite{hochreiter1997long} sequence-to-sequence models for inflection. In \texttt{IMS-CUB-1}, the generation component is based on the architecture by \citet{bahdanau2015neural}, but with fewer parameters, as suggested by \citet{kann-schutze-2016-single}. This model -- as well as all other inflection components used for systems in this category -- receives the sequence of the lemma's characters and the paradigm slot number as input and produces a sequence of output characters.

Their second system, \texttt{IMS-CUB-2}, uses an LSTM pointer-generator network \cite{see-etal-2017-get} instead. This architecture has originally been proposed for low-resource morphological inflection by \citet{sharma-etal-2018-iit}.

The \textbf{NYU--CUBoulder team} also substituted the baseline's generation component. Their morphological inflection models are ensembles of different combinations of transformer sequence-to-sequence models \cite{DBLP:conf/nips/VaswaniSPUJGKP17} and pointer-generator transformers, a model they introduced for the task.

\texttt{NYU-CUB-1} is an ensemble of 6 pointer-generator transformers, while \texttt{NYU-CUB-2} is an ensemble of 6 vanilla transformers. 
Their 
last system, \texttt{NYU-CUB-3}, 
is an ensemble of all 12 models.

\subsection{Submitted Systems: Segment+Conquer}
The \textbf{KU--CST team} did not modify the baseline directly, but, nevertheless, was 
heavily inspired by it.  Their system first employs a character-segmentation algorithm
to identify stem--suffix splits in both the provided lemma list and the corpus, thus
identifying potential suffix-replacement rules. Next, k-means is used to
cluster the extracted suffixes into allomorphic groups.  These suffixes are then
concatenated with the most frequent stems obtained from the lemma list, and
scored by a language model, in order to arrive at plausible inflectional candidates. This approach is \texttt{KU-CST-2}.

However, \texttt{KU-CST-2} often produces very small inflectional paradigms; unsurprisingly, given that the provided corpora are small as well, and, thus, any particular lemma is only inflected in limited ways -- if at all. Therefore, \texttt{KU-CST-1} expands the lemma list with
a logistic-regression classifier that identifies novel verbs to be added. 

\begin{table*}[h]
    \centering
    \small
\setlength{\tabcolsep}{1.8pt}
    \begin{tabular}{l || r r | r r | r r | r r r}
    \toprule
 & \multicolumn{2}{c|}{\texttt{Baseline}} &	\multicolumn{2}{c|}{\texttt{KU-CST}}	& \multicolumn{2}{c|}{\texttt{IMS-CUB}} & \multicolumn{3}{c}{\texttt{NYU-CUB}}	\\
& \multicolumn{1}{c}{1} & \multicolumn{1}{c|}{2} & \multicolumn{1}{c}{1} & \multicolumn{1}{c|}{2} & \multicolumn{1}{c}{1} & \multicolumn{1}{c|}{2} & \multicolumn{1}{c}{1} & \multicolumn{1}{c}{2} & \multicolumn{1}{c}{3} \\
\midrule
MLT & 9.12\phantom{0}\phantom{0}(17) & \textbf{20.00}\phantom{0}\phantom{0}(17) & 0.22\phantom{0}(254) & 1.30\phantom{0}\phantom{0}\phantom{0}(2) & 14.41\phantom{0}\phantom{0}(17) & 17.35\phantom{0}\phantom{0}(17) & 15.29\phantom{0}\phantom{0}(17) & 15.59\phantom{0}\phantom{0}(17) & 15.88\phantom{0}\phantom{0}(17) \\
FAS & \textbf{6.67}\phantom{0}\phantom{0}(31) & 6.54\phantom{0}\phantom{0}(31) & 1.55\phantom{0}\phantom{0}(11) & 0.74\phantom{0}\phantom{0}\phantom{0}(2) & 2.52\phantom{0}\phantom{0}(31) & 2.70\phantom{0}\phantom{0}(31) & 2.76\phantom{0}\phantom{0}(31) & 2.73\phantom{0}\phantom{0}(31) & 2.74\phantom{0}\phantom{0}(31) \\
POR & \textbf{40.39}\phantom{0}\phantom{0}(34) & 39.56\phantom{0}\phantom{0}(34) & 1.09(1104) & 12.75\phantom{0}\phantom{0}(70) & 38.69\phantom{0}\phantom{0}(34) & 39.17\phantom{0}\phantom{0}(34) & 39.93\phantom{0}\phantom{0}(34) & 39.95\phantom{0}\phantom{0}(34) & 40.07\phantom{0}\phantom{0}(34) \\
RUS & 40.68\phantom{0}\phantom{0}(19) & \textbf{41.68}\phantom{0}\phantom{0}(19) & 0.35\phantom{0}(387) & 7.06\phantom{0}\phantom{0}(10) & 38.63\phantom{0}\phantom{0}(19) & 41.11\phantom{0}\phantom{0}(19) & 39.26\phantom{0}\phantom{0}(19) & 40.00\phantom{0}\phantom{0}(19) & 39.74\phantom{0}\phantom{0}(19) \\
SWE & \textbf{45.07}\phantom{0}\phantom{0}(15) & 40.93\phantom{0}\phantom{0}(15) & 0.93\phantom{0}(588) & 22.82\phantom{0}\phantom{0}(17) & 37.60\phantom{0}\phantom{0}(15) & 39.93\phantom{0}\phantom{0}(15) & 39.80\phantom{0}\phantom{0}(15) & 39.93\phantom{0}\phantom{0}(15) & 40.13\phantom{0}\phantom{0}(15) \\
\midrule
avg. & 28.39\phantom{(0000)} & \textbf{29.74}\phantom{(0000)} & 0.83\phantom{(0000)} & 8.93\phantom{(0000)} & 26.37\phantom{(0000)} & 28.05\phantom{(0000)} & 27.41\phantom{(0000)} & 27.64\phantom{(0000)} & 27.71\phantom{(0000)} \\
\midrule
EUS & 0.06\phantom{0}\phantom{0}(30) & 0.06\phantom{0}\phantom{0}(27) & 0.02\phantom{0}\phantom{0}(30) & 0.01\phantom{0}\phantom{0}\phantom{0}(2) & 0.04\phantom{0}\phantom{0}(30) & 0.06\phantom{0}\phantom{0}(30) & 0.05\phantom{0}\phantom{0}(30) & 0.05\phantom{0}\phantom{0}(30) & \textbf{0.07}\phantom{0}\phantom{0}(30) \\
BUL & 28.30\phantom{0}\phantom{0}(35) & 31.69\phantom{0}\phantom{0}(34) & 2.99\phantom{0}(138) & 4.15\phantom{0}\phantom{0}(13) & 27.22\phantom{0}\phantom{0}(35) & \textbf{32.11}\phantom{0}\phantom{0}(35) & 27.69\phantom{0}\phantom{0}(35) & 28.94\phantom{0}\phantom{0}(35) & 27.89\phantom{0}\phantom{0}(35) \\
ENG & 65.60\phantom{0}\phantom{0}\phantom{0}(4) & \textbf{66.20}\phantom{0}\phantom{0}\phantom{0}(4) & 3.53\phantom{0}\phantom{0}(51) & 17.29\phantom{0}\phantom{0}\phantom{0}(7) & 47.80\phantom{0}\phantom{0}\phantom{0}(4) & 61.00\phantom{0}\phantom{0}\phantom{0}(4) & 50.20\phantom{0}\phantom{0}\phantom{0}(4) & 52.80\phantom{0}\phantom{0}\phantom{0}(4) & 51.20\phantom{0}\phantom{0}\phantom{0}(4) \\
FIN & 5.33\phantom{0}\phantom{0}(21) & \textbf{5.50}\phantom{0}\phantom{0}(21) & 0.39(1169) & 2.08\phantom{0}(108) & 4.90\phantom{0}\phantom{0}(21) & 5.38\phantom{0}\phantom{0}(21) & 5.36\phantom{0}\phantom{0}(21) & 5.47\phantom{0}\phantom{0}(21) & 5.35\phantom{0}\phantom{0}(21) \\
DEU & 28.35\phantom{0}\phantom{0}\phantom{0}(9) & \textbf{29.00}\phantom{0}\phantom{0}\phantom{0}(9) & 0.70\phantom{0}(425) & 4.98\phantom{0}\phantom{0}(40) & 24.60\phantom{0}\phantom{0}\phantom{0}(9) & 28.35\phantom{0}\phantom{0}\phantom{0}(9) & 27.30\phantom{0}\phantom{0}\phantom{0}(9) & 27.35\phantom{0}\phantom{0}\phantom{0}(9) & 27.35\phantom{0}\phantom{0}\phantom{0}(9) \\
KAN & 15.49\phantom{0}(172) & 15.12\phantom{0}(172) & 4.27\phantom{0}\phantom{0}(44) & 1.69\phantom{0}\phantom{0}\phantom{0}(1) & 10.50\phantom{0}(172) & \textbf{15.65}\phantom{0}(172) & 11.10\phantom{0}(172) & 11.16\phantom{0}(172) & 11.10\phantom{0}(172) \\
NAV & 3.23\phantom{0}\phantom{0}\phantom{0}(3) & \textbf{3.27}\phantom{0}\phantom{0}\phantom{0}(3) & 0.13\phantom{0}\phantom{0}(38) & 0.20\phantom{0}\phantom{0}\phantom{0}(2) & 0.33\phantom{0}\phantom{0}\phantom{0}(3) & 1.17\phantom{0}\phantom{0}\phantom{0}(3) & 0.40\phantom{0}\phantom{0}\phantom{0}(3) & 0.43\phantom{0}\phantom{0}\phantom{0}(3) & 0.43\phantom{0}\phantom{0}\phantom{0}(3) \\
SPA & 22.96\phantom{0}\phantom{0}(29) & \textbf{23.67}\phantom{0}\phantom{0}(29) & 3.52\phantom{0}(225) & 10.84\phantom{0}\phantom{0}(40) & 19.50\phantom{0}\phantom{0}(29) & 22.34\phantom{0}\phantom{0}(29) & 20.39\phantom{0}\phantom{0}(29) & 20.56\phantom{0}\phantom{0}(29) & 20.30\phantom{0}\phantom{0}(29) \\
TUR & 14.21\phantom{0}(104) & \textbf{15.53}\phantom{0}(104) & 0.11(1772) & 0.71\phantom{0}(502) & 13.54\phantom{0}(104) & 14.73\phantom{0}(104) & 14.88\phantom{0}(104) & 15.39\phantom{0}(104) & 15.13\phantom{0}(104) \\
\midrule
avg. & 20.39\phantom{(0000)} & \textbf{21.12}\phantom{(0000)} & 1.74\phantom{(0000)} & 4.66\phantom{(0000)} & 16.49\phantom{(0000)} & 20.09\phantom{(0000)} & 17.49\phantom{(0000)} & 18.02\phantom{(0000)} & 17.65\phantom{(0000)} \\
\bottomrule
    \end{tabular}
    \caption{BMAcc in percentages and the number of predicted paradigm slots after merging for all submitted systems and the baselines on all development (top) and test languages (bottom). Best scores are in bold. 
    }
    \label{tab:results}
\end{table*}
\section{Results and Analysis}
\subsection{Results on Development Languages}
To encourage reproducibility, we first report the performance of all systems
on the development languages in the upper part of \autoref{tab:results}.
Although participants were not evaluated on these languages, the results provide
insight and enable future researchers to benchmark their progress, while 
maintaining the held-out status of the test languages.

\subsection{Official Shared Task Results}
We show the official test results in the lower part of  \autoref{tab:results}. \texttt{Baseline-2} obtained the highest BMAcc on average, followed in order by \texttt{Baseline-1}, \texttt{IMS-CUB-2}, and \texttt{NU-CUB-2}. Overall, systems built on top of the baseline, i.e., systems from Retrieval+X, performed better than systems from Segment+Conquer: the best Segment+Conquer system only reached $4.66\%$ BMAcc on average. This shows the effectiveness of the baseline. However, it also shows that we still have substantial room for improvement on unsupervised morphological paradigm completion.

Looking at individual languages, \texttt{Baseline-2} performed best for all languages except for EUS, where \texttt{NYU-CUB-3} obtained the highest BMAcc, and BUL and KAN, where \texttt{IMS-CUB-2} was best.

\subsection{Analysis: Seen and Unseen Lemmas}
We further look separately at the results for lemmas which appear in the corpus and those that do not. While seeing a lemma in context might help some systems, we additionally assume that inflections of attested lemmas are also more likely to appear in the corpus. Thus, we expect the performance for seen lemmas to be higher on average.

Examining the performance with respect to observed \textit{inflected forms} might give cleaner results. However, we instead perform this analysis on a per-lemma basis, since the lemmas are part of a system's \textit{input}, while the inflected forms are not.

\autoref{tab:analysis} shows the performance of all systems for seen and unseen lemmas.
Surprisingly, both versions of the baseline show similar BMAcc for both settings with a maximum difference of $0.12\%$ on average. However, the baseline is the only system that performs equally well for unseen lemmas; \texttt{IMS-CUB-1} observes the largest difference, with an absolute drop of $7.85\%$ BMAcc when generating the paradigms of unseen lemmas. Investigating the cause for \texttt{IMS-CUB-1}'s low BMAcc, we manually inspected the English output files, and found that, for unseen lemmas, many generations are nonsensical (e.g., \textit{demoates} as an inflected form of \textit{demodulate}). This does not happen in the case of seen lemmas. A similar effect has been found by \citet{kann-schutze-2018-neural}, who concluded that this might be caused by the LSTM sequence-to-sequence model not having seen similar character sequences during training. The fact that \texttt{IMS-CUB-2}, which uses another inflection model, performs better for unseen lemmas confirms this suspicion. Thus, additional training of the inflection component of \texttt{IMS-CUB-1} on words from the corpus might improve generation. 
Conversely, the baseline -- which benefits from inflection models specifically catered to low-resource settings -- is better suited to inflecting unseen 
lemmas.
Overall, we conclude that there is little evidence that the difficulty of the task increases for unseen lemmas.  Rather, inflection systems need to compensate for the low contextual variety in their training data.

\begin{table*}[h]
    \centering
    \small
\setlength{\tabcolsep}{1.8pt}
    \begin{tabular}{l || r r | r r | r r | r r r}
    \toprule
& \multicolumn{2}{c|}{\texttt{Baseline}} &	\multicolumn{2}{c|}{\texttt{KU-CST}}	& \multicolumn{2}{c|}{\texttt{IMS-CUB}} & \multicolumn{3}{c}{\texttt{NYU-CUB}}	\\
& \multicolumn{1}{c}{1} & \multicolumn{1}{c|}{2} & \multicolumn{1}{c}{1} & \multicolumn{1}{c|}{2} & \multicolumn{1}{c}{1} & \multicolumn{1}{c|}{2} & \multicolumn{1}{c}{1} & \multicolumn{1}{c}{2} & \multicolumn{1}{c}{3} \\
\midrule
EUS & 0.11\phantom{0}\phantom{0}(30) & 0.11\phantom{0}\phantom{0}(19) & 0.03\phantom{0}\phantom{0}(30) & 0.03\phantom{0}\phantom{0}\phantom{0}(2) & 0.11\phantom{0}\phantom{0}(28) & \textbf{0.19}\phantom{0}\phantom{0}(30) & 0.11\phantom{0}\phantom{0}(30) & 0.11\phantom{0}\phantom{0}(30) & 0.11\phantom{0}\phantom{0}(30) \\
BUL & 25.48\phantom{0}\phantom{0}(35) & 28.93\phantom{0}\phantom{0}(34) & 5.62\phantom{0}(138) & 6.33\phantom{0}\phantom{0}(13) & 27.85\phantom{0}\phantom{0}(35) & 29.70\phantom{0}\phantom{0}(34) & 29.30\phantom{0}\phantom{0}(35) & \textbf{29.78}\phantom{0}\phantom{0}(35) & 29.52\phantom{0}\phantom{0}(35) \\
ENG & 70.80\phantom{0}\phantom{0}\phantom{0}(4) & \textbf{71.20}\phantom{0}\phantom{0}\phantom{0}(4) & 3.02\phantom{0}\phantom{0}(51) & 18.86\phantom{0}\phantom{0}\phantom{0}(7) & 69.60\phantom{0}\phantom{0}\phantom{0}(4) & 70.40\phantom{0}\phantom{0}\phantom{0}(4) & 69.20\phantom{0}\phantom{0}\phantom{0}(4) & 70.00\phantom{0}\phantom{0}\phantom{0}(4) & 70.00\phantom{0}\phantom{0}\phantom{0}(4) \\
FIN & 6.17\phantom{0}\phantom{0}(21) & 6.38\phantom{0}\phantom{0}(21) & 0.70(1169) & 3.60\phantom{0}(108) & 6.11\phantom{0}\phantom{0}(21) & \textbf{6.65}\phantom{0}\phantom{0}(21) & 6.55\phantom{0}\phantom{0}(21) & 6.58\phantom{0}\phantom{0}(21) & 6.57\phantom{0}\phantom{0}(21) \\
DEU & 26.70\phantom{0}\phantom{0}\phantom{0}(9) & 27.00\phantom{0}\phantom{0}\phantom{0}(9) & 1.14\phantom{0}(425) & 8.75\phantom{0}\phantom{0}(40) & 27.40\phantom{0}\phantom{0}\phantom{0}(9) & 27.30\phantom{0}\phantom{0}\phantom{0}(9) & 27.50\phantom{0}\phantom{0}\phantom{0}(9) & \textbf{27.60}\phantom{0}\phantom{0}\phantom{0}(9) & 27.40\phantom{0}\phantom{0}\phantom{0}(9) \\
KAN & 16.35\phantom{0}(171) & 15.61\phantom{0}(172) & 6.61\phantom{0}\phantom{0}(44) & 1.69\phantom{0}\phantom{0}\phantom{0}(1) & 13.99\phantom{0}(172) & \textbf{16.49}\phantom{0}(172) & 14.63\phantom{0}(172) & 14.68\phantom{0}(172) & 14.63\phantom{0}(172) \\
NAV & \textbf{2.96}\phantom{0}\phantom{0}\phantom{0}(3) & \textbf{2.96}\phantom{0}\phantom{0}\phantom{0}(3) & 1.46\phantom{0}\phantom{0}(38) & 2.22\phantom{0}\phantom{0}\phantom{0}(2) & \textbf{2.96}\phantom{0}\phantom{0}\phantom{0}(3) & \textbf{2.96}\phantom{0}\phantom{0}\phantom{0}(3) & \textbf{2.96}\phantom{0}\phantom{0}\phantom{0}(3) & \textbf{2.96}\phantom{0}\phantom{0}\phantom{0}(3) & \textbf{2.96}\phantom{0}\phantom{0}\phantom{0}(3) \\
SPA & 20.97\phantom{0}\phantom{0}(29) & \textbf{21.60}\phantom{0}\phantom{0}(29) & 4.43\phantom{0}(225) & 16.37\phantom{0}\phantom{0}(40) & 20.40\phantom{0}\phantom{0}(29) & 21.14\phantom{0}\phantom{0}(29) & 21.17\phantom{0}\phantom{0}(29) & 21.09\phantom{0}\phantom{0}(29) & 21.14\phantom{0}\phantom{0}(29) \\
TUR & 14.68\phantom{0}(104) & 16.38\phantom{0}(104) & 0.23(1772) & 1.42\phantom{0}(502) & 16.98\phantom{0}(104) & 18.02\phantom{0}(104) & 18.30\phantom{0}(104) & \textbf{18.70}\phantom{0}(104) & 18.50\phantom{0}(104) \\
\midrule
avg. & 20.47\phantom{(0000)} & 21.13\phantom{(0000)} & 2.58\phantom{(0000)} & 6.59\phantom{(0000)} & 20.60\phantom{(0000)} & \textbf{21.43}\phantom{(0000)} & 21.08\phantom{(0000)} & 21.28\phantom{(0000)} & 21.20\phantom{(0000)} \\
\midrule
EUS & 0.06\phantom{0}\phantom{0}(30) & 0.06\phantom{0}\phantom{0}(30) & 0.03\phantom{0}\phantom{0}(30) & 0.00\phantom{0}\phantom{0}\phantom{0}(2) & 0.03\phantom{0}\phantom{0}(30) & 0.04\phantom{0}\phantom{0}(30) & 0.05\phantom{0}\phantom{0}(30) & 0.05\phantom{0}\phantom{0}(30) & \textbf{0.07}\phantom{0}\phantom{0}(30) \\
BUL & 31.11\phantom{0}\phantom{0}(35) & 34.44\phantom{0}\phantom{0}(34) & 0.83\phantom{0}(138) & 2.04\phantom{0}\phantom{0}(13) & 26.59\phantom{0}\phantom{0}(35) & \textbf{34.52}\phantom{0}\phantom{0}(35) & 26.07\phantom{0}\phantom{0}(35) & 28.11\phantom{0}\phantom{0}(35) & 26.26\phantom{0}\phantom{0}(35) \\
ENG & 60.40\phantom{0}\phantom{0}\phantom{0}(4) & \textbf{61.20}\phantom{0}\phantom{0}\phantom{0}(4) & 4.12\phantom{0}\phantom{0}(51) & 15.71\phantom{0}\phantom{0}\phantom{0}(7) & 26.00\phantom{0}\phantom{0}\phantom{0}(4) & 51.60\phantom{0}\phantom{0}\phantom{0}(4) & 31.20\phantom{0}\phantom{0}\phantom{0}(4) & 35.60\phantom{0}\phantom{0}\phantom{0}(4) & 32.40\phantom{0}\phantom{0}\phantom{0}(4) \\
FIN & 4.52\phantom{0}\phantom{0}(21) & \textbf{4.62}\phantom{0}\phantom{0}(21) & 0.12(1169) & 0.98\phantom{0}(108) & 3.69\phantom{0}\phantom{0}(21) & 4.11\phantom{0}\phantom{0}(21) & 4.17\phantom{0}\phantom{0}(21) & 4.37\phantom{0}\phantom{0}(21) & 4.13\phantom{0}\phantom{0}(21) \\
DEU & 30.84\phantom{0}\phantom{0}\phantom{0}(9) & \textbf{32.63}\phantom{0}\phantom{0}\phantom{0}(9) & 0.55\phantom{0}(425) & 3.05\phantom{0}\phantom{0}(40) & 22.95\phantom{0}\phantom{0}\phantom{0}(9) & 30.95\phantom{0}\phantom{0}\phantom{0}(9) & 28.74\phantom{0}\phantom{0}\phantom{0}(9) & 28.63\phantom{0}\phantom{0}\phantom{0}(9) & 28.95\phantom{0}\phantom{0}\phantom{0}(9) \\
KAN & 14.64\phantom{0}(172) & 14.55\phantom{0}(172) & 1.88\phantom{0}\phantom{0}(24) & 1.69\phantom{0}\phantom{0}\phantom{0}(1) & 6.72\phantom{0}(172) & \textbf{14.72}\phantom{0}(172) & 7.27\phantom{0}(172) & 7.33\phantom{0}(172) & 7.28\phantom{0}(172) \\
NAV & 3.26\phantom{0}\phantom{0}\phantom{0}(3) & \textbf{3.30}\phantom{0}\phantom{0}\phantom{0}(3) & 0.00\phantom{0}\phantom{0}(38) & 0.00\phantom{0}\phantom{0}\phantom{0}(2) & 0.07\phantom{0}\phantom{0}\phantom{0}(3) & 0.99\phantom{0}\phantom{0}\phantom{0}(3) & 0.15\phantom{0}\phantom{0}\phantom{0}(3) & 0.18\phantom{0}\phantom{0}\phantom{0}(3) & 0.18\phantom{0}\phantom{0}\phantom{0}(3) \\
SPA & 24.94\phantom{0}\phantom{0}(29) & \textbf{25.74}\phantom{0}\phantom{0}(29) & 3.86\phantom{0}(225) & 8.94\phantom{0}\phantom{0}(40) & 18.60\phantom{0}\phantom{0}(29) & 23.54\phantom{0}\phantom{0}(29) & 19.60\phantom{0}\phantom{0}(29) & 20.03\phantom{0}\phantom{0}(29) & 19.46\phantom{0}\phantom{0}(29) \\
TUR & 13.73\phantom{0}(104) & \textbf{14.70}\phantom{0}(104) & 0.00(1757) & 0.00\phantom{0}(500) & 10.12\phantom{0}(104) & 11.47\phantom{0}(104) & 11.48\phantom{0}(104) & 12.08\phantom{0}(104) & 11.77\phantom{0}(104) \\
\midrule
avg. & 20.39\phantom{(0000)} & \textbf{21.25}\phantom{(0000)} & 1.27\phantom{(0000)} & 3.60\phantom{(0000)} & 12.75\phantom{(0000)} & 19.10\phantom{(0000)} & 14.30\phantom{(0000)} & 15.15\phantom{(0000)} & 14.50\phantom{(0000)} \\
\bottomrule

    \end{tabular}
    \caption{BMAcc in percentages and the number of predicted paradigm slots after merging for all submitted systems and the baselines on all test languages; listed separately for lemmas which appear in the corpus (top) and lemmas which do not (bottom). Best scores are in bold.}
    \label{tab:analysis}
\end{table*}

\section{Where from and Where to?}
\subsection{Previous Work}
Prior to this shared task, 
most research on unsupervised systems for morphology was concerned with developing approaches to segment words into morphemes, i.e., their smallest meaning-bearing units \cite{goldsmith-2001-unsupervised,creutz-2003-unsupervised,creutz2007unsupervised,snyder-barzilay-2008-unsupervised,GOLDWATER200921,kurimo-etal-2010-morpho,kudo-richardson-2018-sentencepiece}.  
These methods were built around the observation that inflectional morphemes are 
very common across word types, and leveraged probability estimates such as maximum likelihood (MLE) or maximum a posteriori (MAP) estimations to determine
segmentation points, or minimum description length (MDL)-based approaches. However, they tended to make assumptions regarding how
morphemes are combined, and worked best for purely concatenative morphology.
Furthermore, these methods had no productive method of handling allomorphy---%
morphemic variance was simply treated as separate morphemes.

The task of unsupervised morphological paradigm completion concerns more than just segmentation: 
besides capturing how morphology is reflected in the word form, it also requires correctly clustering transformations into paradigm slots and, finally, generation of unobserved forms. 

While \newcite{xu-etal-2018-unsupervised-morphology} did discover something similar to paradigms, those paradigms were a means to a segmentation end and the shape or size of the paradigms was not a subject of their research. \citet{moon-etal-2009-unsupervised} similarly uses segmentation and clustering of affixes to group words into \emph{conflation sets}, groups of morphologically related words, in an unsupervised way. Their work assumes prefixing and suffixing morphology. In a more task-driven line of research, \newcite{soricut-och-2015-unsupervised} develop an approach to learn morphological transformation rules from observing how consistently word embeddings change between related word forms, with the goal of providing useful word embeddings for unseen words. 

Our task further differs from traditional paradigm completion \citep[e.g.,][]{dreyer-eisner-2011-discovering,ahlberg-etal-2015-paradigm} in that \emph{no} seed paradigms are observed. Thus, no information is being provided regarding the paradigm size, inflectional features, or relationships between lemmas and inflected forms.
Other recent work \citep{nicolai-yarowsky-2019-learning,nicolai-EtAl:2020:LREC} learned fine-grained morphosyntactic tools from the Bible, though they leveraged supervision projected from higher-resource languages \citep{yarowsky-etal-2001-inducing,tackstrom-etal-2013-token}.

\paragraph{Past shared tasks. }

This task extends a tradition of SIGMORPHON shared tasks concentrating on 
inflectional morphology.

The first such task \cite{cotterell-etal-2016-sigmorphon} encouraged participants to
create inflectional tools in a typologically diverse group of 10 languages.
The task was fully-supervised, requiring systems to learn inflectional
morphology from a large annotated database.  This task is similar to human
learners needing to generate inflections of previously unencountered 
word forms, after having studied thousands of other types.

The second task \cite{cotterell-etal-2017-conll} extended the first task from
10 to 52 languages and started to encourage the development of tools for the low-resource setting. While the first shared task approximated an adult learner with experience with thousands
of word forms, low-resource inflection was closer to the language learner that
has only studied a small number of inflections---however,
it was closer to L2 learning than L1, as it still required training sets
with lemma--inflection--slot triplets. The 2017 edition of the shared task also introduced a paradigm-completion subtask:
participants were given partially observed paradigms and asked to generate missing forms, based on complete paradigms observed during training.  This could be described as the supervised version of our unsupervised task, and notably did not require participants to 
identify inflected forms from raw text---a crucial step in L1 learning.

The third year of the shared task \cite{cotterell-etal-2018-conll} saw a further 
extension to more than 100 languages and another step away from supervised
learning, in the form of a contextual prediction task.  This task stripped
away inflectional annotations, requiring participants to generate an
inflection solely utilizing a provided lemma and sentential cues.  This
task further imitated language learners, but extended beyond morphological 
learning to morphosyntactic incorporation.  Furthermore, 
removing the requirement of an inflectional feature vector more closely 
approximated the 
generation step in our task.  However, it was still supervised in that
participants were provided with lemma--inflection pairs in context during training.
We, in contrast, made no assumption of the existence of such pairs.

Finally, the fourth iteration of the task \cite{mccarthy-etal-2019-sigmorphon} 
again concentrated on
less-supervised inflection.  Cross-lingual training allowed low-resource 
inflectors to leverage information from high-resource languages, while
a contextual analysis task flipped the previous year's contextual task on its
head---tagging a sentence with inflectional information.  This process is very
similar to the retrieval portion of our task. We extended this effort to not
only identify the paradigm slot of particular word, but to combine learned
information from each class to extend and complete existing paradigms.
Furthermore, we lifted the requirement of named inflectional features, more closely
approximating the problem as approached by L1 language learners.

\subsection{Future Shared Tasks}
Future editions of the shared task could extend this year's Task 2 to a larger variety of languages or  parts of speech.
Another possible direction is to focus on derivational morphology instead of or in addition to inflectional morphology. 
We are also considering merging Task 2 with the traditional morphological inflection task: participants could then choose to work on the overall task or on either of the retrieval or generation subproblem. 

Finally, we are looking into extending the shared task to use speech data as input. This is closer to how L1 learners acquire morphological knowledge, and, while this could make the task harder in some aspects, it could make it easier in others.

\section{Conclusion}
We presented the findings of the 
SIGMORPHON 2020 shared task on unsupervised morphological paradigm completion (SIGMORPHON 2020 Task 2), in which participants were asked to generate paradigms without explicit supervision. 

Surprisingly, no team was able to outperform the provided baseline, a pipeline system, on average over all test languages. Even though 2 submitted systems were better on 3 individual languages, this highlights that the task is still an open challenge for the NLP community. We argue that it is an important one: systems obtaining high performance will be able to aid the development of human language technologies for low-resource languages. 

All teams that participated in the shared task devised modular approaches. 
Thus, it will be easy to include improved components in the future as, for instance, systems for morphological inflection improve. 
We released all data, the baseline, the evaluation script, and the system outputs in the official repository,\footnote{\url{https://github.com/sigmorphon/2020/tree/master/task2}}
in the hope that this shared task will lay the foundation for future research on unsupervised morphological paradigm completion.

\section*{Acknowledgments}
First and foremost, we would like to thank all of our shared task participants. 
We further thank the passionate morphologists who joined for lunch in Florence's \emph{mercato centrale} on the last day of ACL 2019 to plan the 2020 shared task, as well as the SIGMORPHON Exec, who made this shared task possible.

\bibliography{anthology,acl2020}
\bibliographystyle{acl_natbib}

\end{document}